\documentclass[10pt,conference]{IEEEtran}
\IEEEoverridecommandlockouts

\newif\ifarXiv
\arXivtrue

\usepackage{cite}
\usepackage{amsmath,amssymb,amsfonts}
\usepackage{algorithmic}
\usepackage{graphicx}
\usepackage{textcomp}
\usepackage{xcolor}

\usepackage{url}

\ifarXiv
\usepackage{ascmac}
\fi

\def\BibTeX{{\rm B\kern-.05em{\sc i\kern-.025em b}\kern-.08em
    T\kern-.1667em\lower.7ex\hbox{E}\kern-.125emX}}
\begin{document}

\title{Design by Contract Framework for\\Quantum Software\\
\thanks{This research is based on the discussion at Top SE education program.}
}

\author{\IEEEauthorblockN{1\textsuperscript{st} Masaomi Yamaguchi}
\IEEEauthorblockA{\textit{Quantum Laboratory} \\
\textit{Fujitsu Limited}\\
Kanagawa, Japan \\
y.masaomi@fujitsu.com}
\and
\IEEEauthorblockN{2\textsuperscript{nd} Nobukazu Yoshioka}
\IEEEauthorblockA{\textit{Waseda University / National Institute of Informatics} \\
Tokyo, Japan \\
nobukazuy@acm.org}
}

\ifarXiv
    \onecolumn
    \begin{screen}
       ©2023 IEEE. Personal use of this material is permitted. Permission
        from IEEE must be obtained for all other uses, in any current or future
        media, including reprinting/republishing this material for advertising or
        promotional purposes, creating new collective works, for resale or
        redistribution to servers or lists, or reuse of any copyrighted
        component of this work in other works.
    \end{screen}
    \twocolumn
\fi

\maketitle
\makeatletter
\newenvironment{codemath}{%
    \start@align\tw@\st@rredtrue\m@ne%
}{%
    \endalign
}
\makeatother

\newenvironment{tarray}[2][c]{%
\settowidth{\dimen1}{${} = {}$}%
   \setlength{\arraycolsep}{0.125\dimen1}
   \hspace{-0.125\dimen1}\array[#1]{#2}\relax
}
{%
   \endarray\hspace{-0.125\dimen1}
}

\newcommand{\mi}[1]{\ensuremath{\mathit{#1}}}
\newcommand{\mr}[1]{\ensuremath{\mathrm{#1}}}
\newcommand{\ms}[1]{\ensuremath{\mathsf{#1}}}
\newcommand{\mt}[1]{\ensuremath{\mathtt{#1}}}
\newcommand{\mc}[1]{\ensuremath{\mathcal{#1}}}

\newcommand{\key}[1]{\ensuremath{\mathbf{#1}}} 

\newcommand{\bb}{\tarray{lllll}}
\newcommand{\bbc}{\tarray{c}}
\newcommand{\bbt}{\tarray[t]{lllll}}
\newcommand{\bbb}{\tarray[b]{lllll}}
\newcommand{\ee}{\endtarray}

\newcommand{\A}{\;}
\newcommand{\con}[1]{\ms{#1}}
\newcommand{\var}[1]{\mi{#1}}

\newcommand{\bra}[1]{\left\langle{#1}\right\rvert}
\newcommand{\ket}[1]{\left\lvert{#1}\right\rangle}
\newcommand{\braket}[2]{\left\langle{#1}\middle|{#2}\right\rangle}
\newcommand{\expect}[3]{\left\langle{#1}\middle|{#2}\middle|{#3}\right\rangle}

\newcommand{\up}{\string^}
\newcommand{\til}{\string~}

\newif\ifdraft
\drafttrue

\ifdraft 
\definecolor{darkpastelgreen}{rgb}{0.01, 0.75, 0.24}
\newcommand{\revised}[1] {%
  \textcolor{darkpastelgreen}{#1}%
}
\newenvironment{revisedblock}{%
  \color{darkpastelgreen}%
}{%
  \ignorespacesafterend%
}
\else
\definecolor{darkpastelgreen}{rgb}{0.01, 0.75, 0.24}
\newcommand{\revised}[1] {%
  \textcolor{black}{#1}%
}
\newenvironment{revisedblock}{%
  \color{black}%
}{%
  \ignorespacesafterend%
}
\fi
\begin{abstract}
To realize reliable quantum software, techniques to automatically ensure the quantum software's correctness have recently been investigated.
However, they primarily focus on fixed quantum circuits rather than the procedure of building quantum circuits.
Despite being a common approach, the correctness of building circuits using different parameters following the same procedure is not guaranteed.
To this end, we propose a design-by-contract framework for quantum software.
Our framework provides a python-embedded language to write assertions on the input and output states of \emph{all quantum circuits built by certain procedures}. 
Additionally, it provides a method to write assertions about the statistical processing of measurement results to ensure the procedure's correctness for obtaining the final result.
These assertions are automatically checked using a quantum computer simulator.
For evaluation, we implemented our framework and wrote assertions for some widely used quantum algorithms.
Consequently, we found that our framework has sufficient expressive power to 
verify the whole procedure of quantum software.
\end{abstract}


\section{Introduction}
\label{sec:introduction}
\emph{Quantum computing} is an emerging technology that accelerates many computing-heavy tasks.
Recently, software development frameworks for quantum software, such as Qiskit \cite{qiskit}, have been developed.
However, realizing reliable quantum software is still challenging because of the property of quantum computing:
\emph{superposition},  \emph{entanglement} and \emph{interference}.

For those challenges, different approaches have been studied,
such as formal verification \cite{Hietala:SQIR,Paykin:qwire,Chareton2021:Qbricks},
checking assertions on quantum computers \cite{huang2019,li2020},
equivalence checking of circuits \cite{Burgholzer:eq_check}, 
and automatic testing \cite{honarvar2020,wang2021}.
However, these techniques mainly focus on fixed quantum circuits,
so the correctness of building circuits using different parameters following the same procedure is not guaranteed despite being a common practice.

In this study, we propose a design-by-contract framework \cite{meyr1986} for quantum software.
Our contributions are as follows:
\begin{itemize}
      \item We present a python-embedded language to write assertions about the input states (termed \emph{pre-state}) and output states (termed \emph{post-state}) of \emph{all} quantum circuits built by certain procedures. This enables us to verify the procedure of constructing quantum circuits.
      Additionally, one can write assertions about the statistical processing of measurement results for obtaining the final result.
      These assertions are checked on a quantum computer simulator. 
      \item We provide a module-like feature to construct a bigger circuit from smaller circuits verified by assertions.
\end{itemize}

\section{Motivating Example}
\label{sec:motivating_example}

\begin{figure}[t]
\centerline{\includegraphics[scale=0.522]{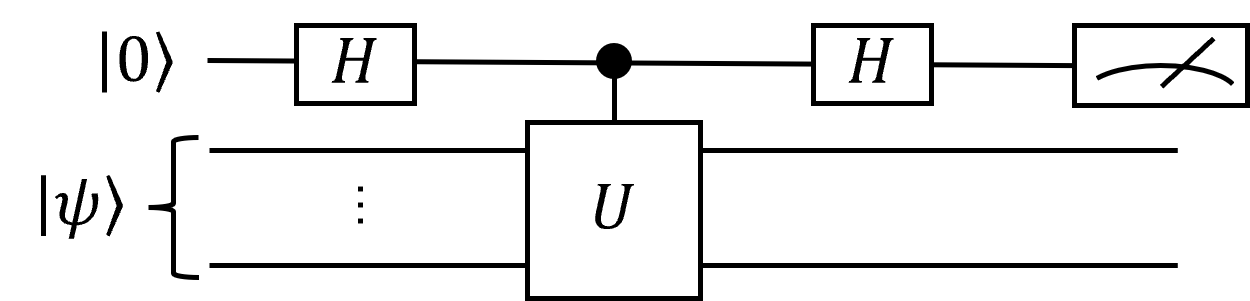}}
\caption{The circuit for the Hadamard test}
\label{fig:hadamard_test}
\end{figure}

As a motivating example, we use the \emph{Hadamard test}. The circuit is shown in Fig. \ref{fig:hadamard_test}.
It is a basic algorithm to estimate the expected value of a unitary matrix $U$ for a state $\ket{\psi}$, i.e., estimate $\expect{\psi}{U}{\psi}$.
It changes the qubits' state as follows:
\begin{equation}
\ket{\psi} \otimes \ket{0} \rightarrow \frac{1}{2}\{(I+U)\ket{\psi} \otimes \ket{0} + (I-U)\ket{\psi} \otimes \ket{1}\}
\label{eq:hadamard_test_circuit}
\end{equation}
Thus, the real part of $\expect{\psi}{U}{\psi}$ is obtained by:
\begin{equation}
 \mathrm{Re} \expect{\psi}{U}{\psi} =  p_0 - p_1 \approx \frac{N_0 - N_1}{N_0+N_1}
\label{eq:hadamard_test_measure_result}
\end{equation}
where $p_i$ and $N_i$ are the probability and count of obtaining $\ket{i}$ when measuring the first qubit, respectively. 

The circuit adopts $U$ as a parameter.
Thus, 
we need to verify \emph{the procedure for building the circuit from the gate $U$}.
This includes complex steps such as the decomposition of gates to run on quantum computers.
Moreover, we need statistical postprocessing according to \eqref{eq:hadamard_test_measure_result} to obtain the final result.
Unfortunately, existing works cannot verify these procedures.


\section{Proposed Approach}
\label{sec:proposed_approach}
Our approach is based on the design-by-contract framework \cite{meyr1986}.
We provide a language leveraging Qiskit to write assertions about pre/post-states of all circuits built by certain procedures and the statistical processing of measurement results.


\subsection{Verifying the Procedure of Building Quantum Circuits}

\begin{figure}[t]
\centering
{\scriptsize
\vskip-2ex
\begin{codemath}
\bb
\key{def}~\var{make\_circ\_hadamard\_test}(\var{Ugate},\var{U})\rightarrow \con{AssertQuantumCircuit}:\\
    \qquad \bbt
    \var{circ} = \con{AssertQuantumCircuit}(\var{size}=\var{Ugate.num\_qubits} + 1)\\
    \var{circ}.\var{append}(\con{HGate}(), [0])\\
    \var{ctrl\_U} = \var{decompose}(\bbt \var{Ugate}.\var{control}(), \\
    \var{basis\_gates}=[\mt{"h"},\mt{"rx"},\mt{"rz"},\mt{"cx"}])\ee\\
    \var{circ}.\var{append}(\bbt \var{ctrl\_U}, \var{range}(\var{Ugate}.\var{num\_qubits} + 1))\ee\\
    \var{circ}.\var{append}(\con{HGate}(), [0])\\
    \\
    \key{def}~\var{condition}(\var{pre\_state}, \var{post\_state})
    \rightarrow \con{bool}:\\
    \qquad \bbt
        \var{psi} = \con{StateFn}(\var{partial\_state}(\bbt\var{pre\_state},\\
                    \var{range}(1, \var{Ugate}.\var{num\_qubits} + 1)))\ee\\
        \var{state0} = ((\var{psi} + (\var{U} @ \var{psi})) / 2)\up\con{Zero}\\
        \var{state1} = ((\var{psi} - (\var{U} @ \var{psi})) / 2)\up\con{One}\\
        \key{return}~\var{eq\_state}(\var{post\_state}, \var{state0} + \var{state1})\ee\\
        \\
        \var{circ}.\var{add\_condition}(\mt{"condition1"}, \var{condition})\\
    \key{return}~\var{circ}\\
    \ee
\ee
\end{codemath}
}
 \caption{Code with assertions for Hadamard test written in our framework}
 \label{fig:code_hadamard_test_circuit}
\end{figure}
Fig. \ref{fig:code_hadamard_test_circuit} is an example code for the Hadamard test written in our framework. 
It takes the gate $\var{Ugate}$ and its unitary representation $U$ and returns a circuit with an assertion for \eqref{eq:hadamard_test_circuit}. 
This code represents the procedure of building the circuit from the gate $Ugate$.
It consists of three parts: define a circuit including a decomposing step, define an assertion, and return the circuit with the assertion tagged \mt{"condition1"}.

\subsection{Checking the Correctness of Statistical Postprocessing}

In quantum computing, we must measure qubits and statistically process the measurement result to obtain the final result.
Our framework can verify this process, as shown in Fig. \ref{fig:code_run_hadamard_test_circuit_measure}.

\begin{figure}[t]
\centering
{\scriptsize
\begin{codemath}
\bb
    \var{U} = \con{PrimitiveOp}(\con{Operator}([[1, 0], [0, \var{cmath}.\var{exp}(1\mathrm{j} * \var{math}.\var{pi} / 4)]])
    )\\
    \var{Ugate} = \con{TGate}()\\
    \var{circ\_hadamard\_test} = \var{make\_circ\_hadamard\_test}(\var{Ugate}, \var{U})\\
    \\
    \var{psi} = \con{Plus}\\
    \var{circ} = \con{AssertQuantumCircuit}(\var{size} = 2)\\
    \var{circ}.\var{append}(\con{HGate}(), [1])\\
    \var{circ}.\var{append}(\var{circ\_hadamard\_test}, [0, 1])\\
    \\
    \key{def}~\var{estimate\_exp}(\var{result}) \rightarrow \con{float}:\\
    \qquad \bbt
        \var{c} = \var{result}.\var{get\_counts}()\\
        \key{return}~(\var{c}[\mt{"0"}] - \var{c}[\mt{"1"}]) / (\var{c}[\mt{"0"}] + \var{c}[\mt{"1"}])
    \ee\\
    \\
    \var{circ\_measure}: \con{AssertQuantumCircuitMeasure}\\
    = \var{circ}.\var{measure}(\var{postprocess}=\var{estimate\_exp}, \var{qubit}=[0])\\
    \\
    \key{def}~\var{measure\_condition}(
        \var{pre\_measure\_state},
        \var{result},
        \var{est\_exp}
    )\rightarrow\con{bool}:\\
    \qquad\bbt
        \var{actual\_exp} = ((\til\var{psi}) @ \var{U} @ \var{psi}).\var{eval}().\var{real}\\
        \key{return}~\var{math}.\var{isclose}(\var{actual\_exp}, \var{est\_exp}, \var{abs\_tol}=0.01)
        \ee\\
    \\
    \var{circ\_measure}.\var{add\_condition}(\mt{"condition2"}, \var{measure\_condition})\\
    \var{print}(\var{circ\_measure}.\var{run}(\var{shots}=100\_000))
    \\
    \\
    \texttt{>>}~\texttt{0.85514}\\
\ee&&
\end{codemath}
}
 \caption{Code to check the post process of the Hadamard test}
 \label{fig:code_run_hadamard_test_circuit_measure}
\end{figure}

This code estimates $\mr{Re}\expect{+}{T}{+}$.
Here, we defined an assertion to check whether statistical processing follows \eqref{eq:hadamard_test_measure_result}.
The assertion's arguments are the qubits' state before measurement, result of measurement, and result of post-processing.
Note that we separated the class of the measurement and the other steps because the former contains classical functions. 
We built the circuit from smaller circuits by using our module-like system.
The assertions in nested circuits are verified in runtime.

By running the code, it prints the estimated value without any assertion error, 
which means that the whole procedure of the Hadamard test is correct with respect to $T$ and $\ket{+}$.

If the circuit does not behave as expected, our framework raises an exception to notify the user, as shown in Fig. \ref{fig:code_run_hadamard_test_circuit_wrong}.

\begin{figure}[t]
\centering
{\tiny
\begin{verbatim}
Cell In[8], line 22
---> 22 print(circ_measure.run(shots=100_000))
...
StateConditionError: Condition Error occurred in 'condition1'
\end{verbatim}
}
 \caption{Exception when the assertion of the Hadamard test fails}
 \label{fig:code_run_hadamard_test_circuit_wrong}
\end{figure}

\section{Evaluation}
\label{sec:evaluation}

We evaluated our framework by writing three quantum algorithms that are widely used: the Hadamard test, quantum Fourier transform (QFT), and quantum phase estimation (QPE). The details of the evaluation are not provided for the sake of brevity.
As explained in section \ref{sec:proposed_approach}, our framework can express the whole procedure of quantum software and rapidly detects mistakes through the failure of the assertions. 

The limitation of our framework is the lack of intermediate measurements. 
We only allow measurements in the last of the circuit, but some cases, such as error correction, need it.
\section{Conclusion and Future Direction}
\label{sec:conclusion}

We proposed a design-by-contract framework for quantum software, whose novelty is that it can verify the procedure of building quantum circuits and statistical post-processing.
We demonstrated the effectiveness of our framework and clarified its limitations.
In future studies, we will expand our method to handle intermediate measurements.
Another direction is to evaluate the effectiveness of our method using more examples.

\bibliographystyle{IEEEtran}
\bibliography{IEEEabrv,reference}

\end{document}